%% file: main.tex
\documentclass[10pt,twocolumn,letterpaper]{article}

\usepackage[pagenumbers]{cvpr} 

\input{preamble}

\definecolor{cvprblue}{rgb}{0.21,0.49,0.74}
\usepackage[pagebackref,breaklinks,colorlinks,citecolor=cvprblue]{hyperref}
\usepackage[accsupp]{axessibility}
\usepackage{moresize}

\author{Vasileios Baltatzis$^1$, Rolandos Alexandros Potamias$^1$, Evangelos Ververas$^1$, \\Guanxiong Sun$^2$, Jiankang Deng$^1$, Stefanos Zafeiriou$^1$ \\
\\
$^1$Imperial College London, $^2$Queen’s University Belfast\\
{\tt\small \{{vasileios.baltatzis18, r.potamias, e.ververas16, j.deng16, s.zafeiriou}\}@imperial.ac.uk}, \\
{\tt\small gsun02@qub.ac.uk}
}

\begin{document}

\title{Neural Sign Actors: A diffusion model for 3D sign language production from text}

\twocolumn[{
\renewcommand\twocolumn[1][]{#1}%
\maketitle
\begin{center}
    \centering
    \captionsetup{type=figure}
    \includegraphics[width=0.95\textwidth]{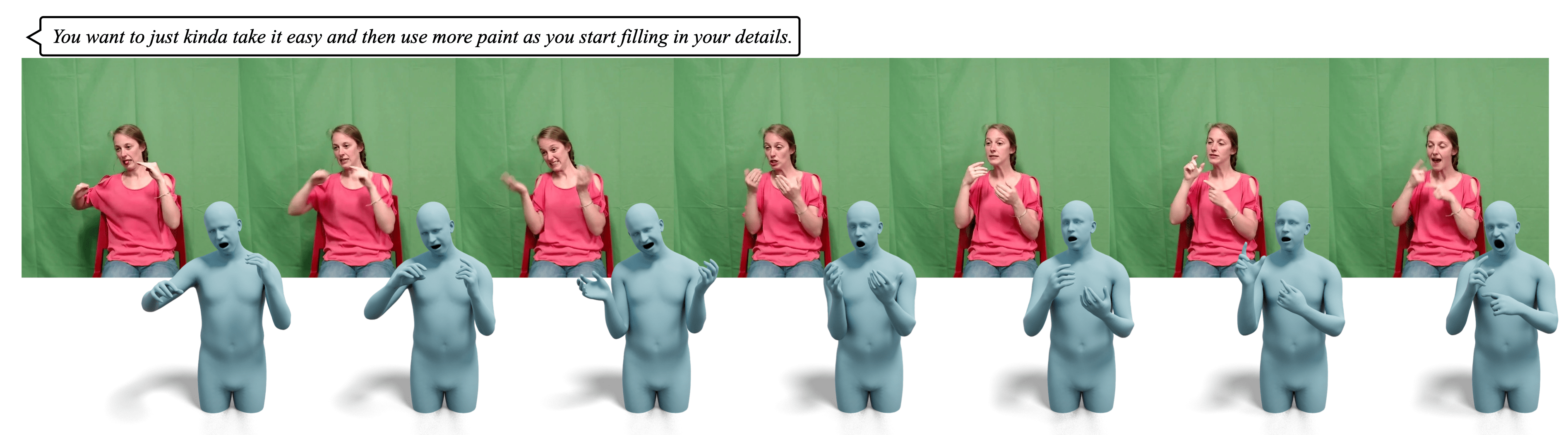}
    \captionof{figure}{The proposed method takes raw text as input and generates a realistic and coherent motion of its corresponding sign language translation. From top to bottom: the input text, the ground truth sign language video (shown just for reference), and the generated motion.}
\end{center}}]

\maketitle
\input{sec/0_abstract}    
\input{sec/1_intro}
\input{sec/2_related_work}
\input{sec/3_dataset}
\input{sec/4_method}
\input{sec/5_experiments}
\input{sec/6_conclusion}
{
    \small
    \bibliographystyle{ieeenat_fullname}
    \bibliography{main}
}


\end{document}

%% file: preamble.tex
\usepackage[dvipsnames]{xcolor}


%% file: sec/0_abstract.tex
\begin{abstract}
Sign Languages (SL) serve as the primary mode of communication for the Deaf and Hard of Hearing communities. Deep learning methods for SL recognition and translation have achieved promising results. However, Sign Language Production (SLP) poses a challenge as the generated motions must be realistic and have precise semantic meaning. Most SLP methods rely on 2D data, which hinders their realism. In this work, a diffusion-based SLP model is trained on a curated large-scale dataset of 4D signing avatars and their corresponding text transcripts. The proposed method can generate dynamic sequences of 3D avatars from an unconstrained domain of discourse using a diffusion process formed on a novel and anatomically informed graph neural network defined on the SMPL-X body skeleton. Through quantitative and qualitative experiments, we show that the proposed method considerably outperforms previous methods of SLP. This work makes an important step towards realistic neural sign avatars, bridging the communication gap between Deaf and hearing communities.\footnote{Project page: \ssmall \url{https://baltatzisv.github.io/neural-sign-actors/}}
\end{abstract}

%% file: sec/1_intro.tex
\section{Introduction}
\label{sec:intro}

Sign language (SL) is a form of language in which visual-manual modalities are used instead of spoken words to convey meaning. It is the predominant form of communication for more than 70 million Deaf and Hard of Hearing people around the world. Akin to verbal languages, SLs have extremely rich vocabulary and grammar, yet the complexities differ drastically \cite{sutton1999linguistics}. To enable effective visual communication, they consist of both manual and non-manual components \cite{SLT_survey}. The manual modality encompasses hand articulation, orientation, position, and motion, while non-manual elements include arm movements and facial expressions \cite{braem2001hands}. Whilst it is possible to convey some meaning using just hand articulations, expressiveness is limited since non-manual elements often convey emotions \cite{sutton1999linguistics,antonakos2015survey}. 

Recently, several methods have been proposed to bridge the domain gap between sign and spoken languages. Most methods focus on Sign Language Recognition (SLR) which includes the translation of a specific sign to its corresponding meaning, as well as Sign Language Translation (SLT) that extends SLR to the translation of a sign sequence to its spoken word equivalent. This is usually tackled using glosses \cite{camgoz2020sign,hao2021self,chen2022simple,chen2022two}, which are simplified mid-level representations that relate each sign with a corresponding meaning. However, even though glosses have provided a substantial enhancement to SLT methods, they have a predefined informative bottleneck, which limits the translation accuracy, and they usually fail to provide long-range dependencies and contextual information~\cite{camgoz2020sign,kapoor2021towards,zhou2023gloss}. 

Despite the significant number of individuals with hearing difficulties, only $\sim$5\% of television programs are interpreted into sign language, which shows the vital need for 3D signing avatars. Compared to SLR and SLT, only a small number of methods have attempted to tackle the task of Sign Language Production (SLP), either by using directly stand-alone glosses \cite{stoll2018sign,zelinka2019nn} or by training a network to map text to glosses \cite{saunders2021mixed}. In the SLP setting, a network is given a text sentence and attempts to generate a motion that reflects the corresponding sign language translation. Usually, this is done using 2D and 3D joints to represent the human body~\cite{stoll2018sign,kapoor2021towards,saunders2020progressive}. However, joints provide an unrealistic representation of the animation, limiting their practical use in real-world avatar actor applications. Recently, Stoll \etal \cite{stoll2022there} proposed to extend SLP to 3D meshes using an optimization step that fits a SMPL-X \cite{SMPL-X} model to the predicted 2D joints. In contrast, the proposed method directly regresses the poses of SMPL-X \cite{SMPL-X} model to generate an animatable 3D signing avatar.  

Aside from its challenging nature, SLP remains relatively unexplored due to the absence of large-scale available datasets with a sufficient vocabulary size. In particular, the majority of current SLP methods rely on German sign language datasets composed of only a few thousand words \cite{forster2014extensions,camgoz2018neural} and use 2D landmarks detected from off-the-shelf pose estimation methods \cite{OPENPOSE}. In contrast, we employ a hybrid regression-optimization method to accurately annotate, with SMPL-X pose and shape parameters \cite{SMPL-X}, a large-scale video dataset \cite{duarte2021how2sign} composed of over 16k word tokens. Using the acquired 3D pose annotations, we train a dynamic diffusion model to learn SLP from English texts. Our method directly translates text to signs without using any intermediate representation \cite{saunders2020progressive}, which increases the generative capacity of the network. Given that sign language cannot be translated word-for-word \cite{SLT_survey}, we utilize an off-the-shelf sentence encoder \cite{CLIP} which also enables out-of-distribution generalization.
To sum up, the contributions of this study can be summarized as:
\begin{itemize}
    \item We introduce the task of direct 3D signing \textit{avatar} generation from text, without relying on 2D fitting optimizations or any intermediate gloss representations. In this paper, we aim to make a step towards neural sign avatars to aid the Deaf and Hard of Hearing community \cite{naert2020survey}. 
    
    \item We derive the first large-scale 3D dataset of American Sign Language by designing a state-of-the-art pipeline to annotate the How2Sign dataset \cite{duarte2021how2sign}.
    \item We propose a text-conditioned dynamic diffusion model founded on a novel, anatomically inspired graph neural network that facilitates SLP. The proposed model achieves remarkable results that outperform the current state-of-the-art models, by a large margin. 
\end{itemize}

%% file: sec/2_related_work.tex
\section{Related Work}
\label{sec:related_work}

\subsection{Sign Language Production}
Despite nearly two decades of research \cite{cox2002tessa,mcdonald2016automated}, the development of highly effective sign language production methods remains challenging. Stoll \etal \cite{stoll2018sign} proposed the first neural SLP method forming a seq2seq architecture to map text to glosses. To decode poses to 2D joint locations they proposed an empirical lookup table paradigm. To avoid the two-stage generation, Zelinka \etal \cite{zelinka2019nn} utilized OpenPose \cite{OPENPOSE} to extract joint locations from Czech weather forecasting videos and train a network to directly regress the 2D joint poses. Recently, several methods \cite{saunders2020progressive, hwang2021non}
proposed transformer-based architectures to tackle German sign language production \cite{camgoz2020sign}. However, their generations suffer from under-articulation and limited expressiveness in hand and body motion.
Follow-up works attempted to improve the generation quality using adversarial training \cite{saunders2020adversarial}, mixture density networks \cite{saunders2021mixed} and dictionary representations \cite{Saunders_2022_CVPR}. However, most of the aforementioned methods, apart from being contingent to intermediate glosses representation, rely on the regression of 2D/3D joint positions, a process that inherently encounters difficulties in realistically conveying meanings. In an attempt to tackle such limitations, Stoll \etal \cite{stoll2022there} proposed the application of a post-regression SMPL-X \cite{SMPL-X} fitting to lift 2D joints to 3D meshes. On the contrary, we make a step towards realistic signing avatars and propose a diffusion pipeline that directly regresses SMPL-X poses from an unconstrained domain of discourse, without relying on any intermediate representations such as glosses. 

\begin{figure*}[!ht]
    \centering
    \includegraphics[width=0.8\linewidth]{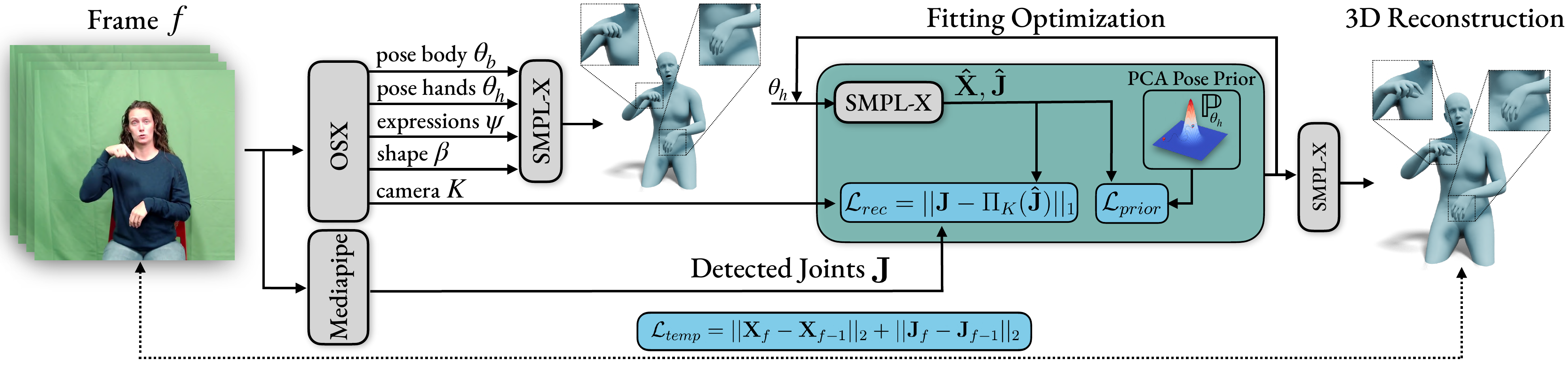}
    \captionof{figure}{\small{\textbf{Overview of the fitting pipeline.} A set of input frames $F$ are first processed by OSX \cite{lin2023one} to obtain an initial set of pose parameters $\mathbf{p}^{init}_{1:F}$. Then, using the Mediapipe algorithm \cite{lugaresi2019mediapipe}, we fine-tune the predicted hand poses to match the detected joints $\mathbf{J}$ while constraining the hand poses $\boldsymbol{\theta_h}$ to lie in the space of plausible poses. Finally, using a temporal coherence loss, we acquire smooth and high-fidelity annotations of 3D signing avatars. }
    \label{fig:fitting}}
\end{figure*}

\subsection{Sign Language Datasets}
A major contributing factor to the slow-paced advancements in sign language research is the absence of large-scale datasets \cite{bragg2019sign}. Earlier datasets were designed with a  focus on sign language recognition using \emph{isolated signs}~\cite{wilbur2006purdue,athitsos2008american,von2008significance,joze2018ms,li2020word}, containing a limited vocabulary. To address the challenges of sign language recognition and translation within the context of complete sentences, several continuous sign language datasets have been introduced. More specifically, RWTH-BOSTON-50 \cite{zahedi2005combination}, Dreuw \etal \cite{dreuw2007speech}, SIGNUM \cite{agris2010signum}, and BSL \cite{schembri2013building} along with DictaSign Corpus, which was developed in several languages \cite{braffort2010sign,efthimiou2010dicta,efthimiou2012dicta}, were among the first datasets with sentence level annotations.
While additional datasets featuring an expanded set of signs have been introduced \cite{chai2014devisign,joze2018ms,li2020word}, it is crucial to emphasize the importance of continuous sign language for the purposes of translation and production. S-pot \cite{viitaniemi-etal-2014-pot} was among the first large-scale continuous sign language datasets with a vocabulary of over 1K signs of Finnish sign language collected in a constrained environment. To enforce the robustness of sign language translation methods, RWTH-Phoenix \cite{camgoz2018neural} contained a collection of TV clips with German sign language, remaining amongst one of the most popular datasets used for sign language translation \cite{camgoz2020sign} and production \cite{saunders2020progressive}. Similarly, BSL-1K, as presented in \cite{albanie2020bsl}, curated a dataset featuring British sign language (BSL) used in casual conversations, encompassing a total vocabulary of 1K signs. Recently, Duarte \etal \cite{duarte2021how2sign} collected How2Sign, a large-scale dataset of American Sign Language (ASL), that is aligned with speech signals from the How2 dataset \cite{sanabria2018how2}. How2Sign is equipped with a vocabulary of over 16K signs, captured in a total of seventy-nine hours of continuous sign language. A major limitation of the above datasets, is the lack of available 3D annotations, that not only aid the translation tasks, but are also essential for training realistic 3D signing avatars. In this work, we have extended the How2Sign dataset by incorporating high-quality SMPL-X \cite{SMPL-X} annotations, thereby establishing it as the first publicly available 3D sign language dataset. 

%% file: sec/3_dataset.tex
\section{Dataset}
\label{sec:fitting}
To train a high-fidelity SLP method, capable of generating realistic sign actors, we curate a large-scale dataset of 3D dynamic ASL sign sequences paired with their corresponding text transcripts. 
To do so, we devise a robust 4D reconstruction pipeline, crafted specifically for hand gestures, to estimate dynamic hand and body poses of signing avatars in the SMPL-X format \cite{SMPL-X}.
How2Sign dataset \cite{duarte2021how2sign} provides the optimum candidate since it is composed of 35K high-resolution clips of \emph{co-articulated} ASL with a substantial vocabulary size featuring over 16K word tokens. 

To acquire high-fidelity 4D reconstructions of How2Sign clips, we build our pipeline upon the powerful OSX \cite{lin2023one}. Specifically, we initialize our fitting optimization using the SMPL-X pose and shape parameters acquired from OSX for each one of the $F$ frames of the clip as:
\begin{equation}
    \mathbf{p}_{1:F}^{init} = [\boldsymbol{\theta_b || \theta_h || \psi || \beta}], 
\end{equation}
where $\boldsymbol{\theta_b, \theta_h, \psi, \beta}$ denote the body pose, hand pose, expression, and shape parameters respectively, and $||$ the concatenation symbol. 

Recognizing that hand poses constitute the pivotal component in conveying SL, we adopt an optimization procedure to enhance the precision of hand poses and rectify any potential misalignments by leveraging body and hand joints detected from the Mediapipe framework \cite{lugaresi2019mediapipe}. More specifically, we optimize the initial pose parameters $\mathbf{p}_{1:F}^{init}$ to minimize the re-projection loss $\mathcal{L}_{rec}$ between the regressed joints $\mathbf{\hat{J}}_{1:F}$ and the joints predicted from Mediapipe $\mathbf{J}_{1:F}$: 
\begin{equation}
    \mathcal{L}_{rec} = ||\mathbf{J}_{1:F} - \Pi_K(\mathbf{\hat{J}}_{1:F})||_1,
\end{equation}
where $\Pi_K$ is the intrinsic camera projection matrix. 
Following extensive experimentation, we observed that optimization is only necessary for the arm and hand joints, as the OSX-regressed body joints are sufficiently accurate. Fitting hand poses using 2D keypoints is an exceptionally challenging task, primarily due to the numerous articulations and the inherent ambiguities within the solutions. While several methods have been proposed to constrain SMPL to feasible body poses \cite{SMPL-X, davydov2022adversarial, tiwari22posendf}, pose prior models for the hand models \cite{romero2022embodied,potamias2023handy} remain unexplored. To constrain the optimization to plausible human and hand poses, we propose a simple but intuitive approach using Principal Component Analysis (PCA) to model the subspace of anatomically feasible poses. Specifically, we trained a PCA pose prior model on two large datasets of human body \cite{AMASS} and hand \cite{fan2023arctic} poses, to model the distribution of feasible arm and hand poses. To formulate the prior loss we measure the reconstruction error of a mesh $\mathbf{X}$ projected and reconstructed from the PCA space $\mathbf{U}$ as: 
\begin{equation}
    \mathcal{L}_{prior} = || \mathbf{X} -  [(\mathbf{X} - \boldsymbol{\mu} )\mathbf{U} ^ T] \mathbf{U} + \boldsymbol{\mu} ||_2,
\end{equation}
where $\mathbf{U} \in \mathbb{R}^{N\cdot3\times d}$ is the eigenvector basis of $d$ components and 
$\boldsymbol{\mu}$ is the mean mesh. 
Intuitively, realistic poses will result in smaller reconstruction errors compared to infeasible articulations. 

Finally, a common issue with blurry videos is that the OSX reconstruction and the Mediapipe detections may include jittering and spatio-temporal noise. To tackle this, we enforce temporal coherence using a loss function on both vertex and joint space that enforces smooth transitions between adjacent frames $f$, $f-1$: 
\begin{equation}
    \mathcal{L}_{temp} = || \mathbf{X}_{f} - \mathbf{X}_{f-1}||_2 + || \mathbf{J}_{f} - \mathbf{J}_{f-1}||_2.
\end{equation}
The overall loss function can be defined as: 
\begin{equation}
    \mathcal{L} = \mathcal{L}_{rec} + \lambda_{prior}\mathcal{L}_{prior} + \lambda_{temp}\mathcal{L}_{temp},
\end{equation}
where $\lambda_{prior}, \lambda_{temp}$ are hyperparameters. An overview of the proposed fitting pipeline is depicted in \cref{fig:fitting}. 

%% file: sec/4_method.tex
\section{Method}

\begin{figure*}[t]
    \centering
    \includegraphics[width=0.85\textwidth]{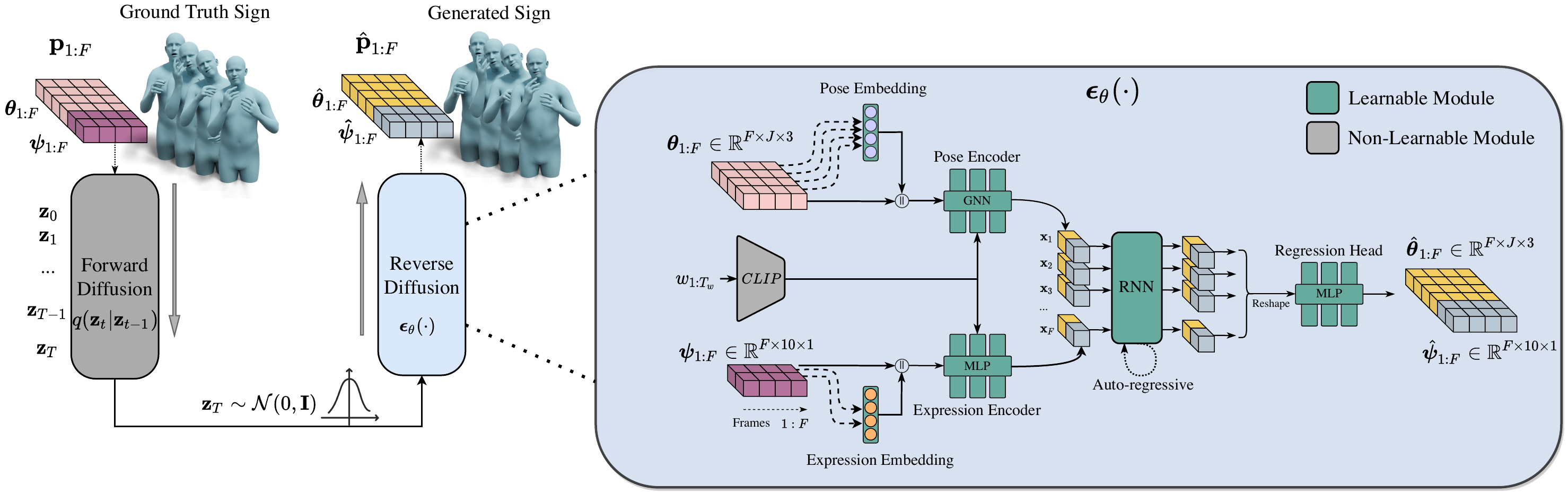}
    \captionof{figure}{\small{\textbf{Overview of the proposed method.} We employ a diffusion model to learn a mapping between text scripts and 3D sign language. The proposed framework consists of an auto-regressive denoising module $\epsilon_\Theta$ that is founded on the novel anatomically informed pose encoder to model the sign motions. }
    \label{fig:method}}
\end{figure*}

We propose Neural Sign Actors, a diffusion-based generative model that generates motion sequences conditioned on text transcripts. Similar to traditional diffusion architectures, our method is composed of the deterministic forward diffusion process that gradually adds noise to the input distributions and the reverse denoising model $\epsilon_\theta(\cdot)$ that predicts the noise introduced by the forward process at each time-step. To reduce the computational requirements and facilitate the generation quality, we train a diffusion model on the low-dimensional pose space defined by SMPL-X \cite{SMPL-X} model instead of the vertex space. Given that sign language is solely related to hand motion and facial expressions, we focus on modeling the pose and the expression parameters on the canonical shape. An overview of the proposed approach can be found in \cref{fig:method}.

\subsection{Forward Diffusion Process}
During the forward diffusion process, noise sampled from a Gaussian distribution $\mathcal{N}(\mu, \sigma\mathbf{I})$ is 
gradually added to the sequence of SMPL-X parameters $\mathbf{p}_{1:F}$, which consist of the concatenated poses $\boldsymbol{\theta}_{1:F}$, and expressions $\boldsymbol{{\psi}}_{1:F}$. This process iterates a total of $T$ times as a Markov chain, ultimately transforming the poses into a Gaussian distribution $\mathcal{N}(\textbf{0}, \mathbf{I})$. In line with the approach delineated in \cite{ho2020denoising}, we establish the forward diffusion process as follows:
\begin{equation}
    q(\mathbf{p}_{1:F}^{t} | \mathbf{p}_{1:F}^{t-1}) =  \mathcal{N}(\mathbf{p}_{1:F}^t | \sqrt{\alpha_t} \mathbf{p}_{1:F}^{t-1}, (1- \alpha_t)\mathbf{I})
\end{equation}
where $\alpha_t$ is the variance schedule parameter that controls the noise scheduling of the process.

\subsection{Reverse Diffusion Process}
Following the forward process, the goal of the denoising module $\epsilon_\Theta$ is to learn the reverse process, \ie learn a mapping from the noised distribution to the real pose space $p_\theta(\mathbf{p}_{1:F}^{t-1} | \mathbf{p}_{1:F}^{t})$. Following the reparameterization trick of \cite{ho2020denoising}, we train a denoising model $\epsilon_\theta$ that predicts the time conditioned noise $\epsilon_t$ as: 
\begin{equation}
    \mathcal{L}_{t} = || \epsilon_t - \epsilon_\Theta (\mathbf{p}_{1:F}^{t}, t, \boldsymbol{w}_{1:F})||_2, 
\end{equation}
where $\epsilon_t$ is the noise added at time-step $t$ of the forward diffusion process and $w_{1:F}$ denotes the target text transcript. To further enforce the generation of accurate hand articulations, we modify $\mathcal{L}_{t}$ to double the weighting factor of the hand poses.   
The proposed denoising module can be divided into three main components: the anatomically informed pose and expression encoders, the text encoder, and the auto-regressive decoder.

\noindent\textbf{Anatomically Informed Encoder.}
\label{sec:gnn}
Previous methods for human motion generation attempted to model poses and joint rotations independently, using permutation equivariant layers such as MLPs \cite{guo2022generating,chen2023executing}. We observed that such equivariance limits the generative ability of the network and results in mild motion intensities. To tackle this limitation, we propose to break the permutation equivariance using a novel, anatomically inspired, graph neural network (GNN) combined with a pose embedding layer. In particular, for a joint $i$, we build a message passing layer that updates the joint $i$ features $\mathbf{f}_i$ based on the relative features of the SMPL-X kinematic tree $\mathcal{K}$. Additionally, to break the permutation equivariance of the proposed message passing layer, we introduce a pose embedding that encodes joint index $i$ into a unique token feature $\mathcal{P}_i$. With this formulation, the network learns to disentangle the joint distributions since each joint is uniquely defined by its token feature $\mathcal{P}$. The update function of the proposed message passing layer can be defined as: 
\begin{equation}
    \mathbf{f}'_i = \gamma\left(\sum_{j \in \mathcal{K}_i} {g}_{ij}(\mathbf{f}_j - \mathbf{f}_i) + \mathcal{P}_i\right),
 \end{equation}
where $\mathbf{f}_j$ denotes the features of joint $j$ which is anatomically connected to joint $i$, in the kinematic tree $\mathcal{K}$, $g_{ij}$ is an anisotropic function between the joints $i,j$, $\mathcal{P}_i$ refers to the positional encoding of joint $i$, and $\gamma$ is a non-linearity. We establish the anisotropy of kernel $g_{ij}$ by assigning a different set of learnable weights to each set of neighbors. 

Similarly, to break the permutation equivariance of the expression encoder layers, we append each of the expression parameters with a learnable expression token $\mathcal{E}$. Given that expression blendshapes cannot be represented in graph form, we utilize an MLP to encode their latent features as: 
\begin{equation}
    \mathbf{g}'_i = \gamma\left(  \text{MLP}\left(\mathbf{g}_i + \mathcal{E}_i \right) \right),
 \end{equation}
where $\mathbf{g}_i$ denotes the latent features of expression parameter $i$ and $\mathcal{E}_i$ refers to its corresponding expression embedding.

\noindent\textbf{Text Encoding.}
Sign language is not merely a direct word-for-word translation of spoken language, rather it possesses its own unique grammar, semantic structure, and distinct language logic \cite{li2020tspnet}. Contingent upon this, we avoid using a sequence of word embeddings to condition the motion generation and propose to utilize CLIP \cite{CLIP} as a powerful sentence encoder that is able to generalize to arbitrary text prompts. We condition pose and expression encoders on the text embedding using a gating approach described in \cite{grathwohl2018ffjord}. 

\noindent \textbf{Auto-regressive Decoder.}
Considering that motion can be conceptualized as a sequence of poses where each pose is contingent upon its predecessor, we constructed our motion generative network utilizing an auto-regressive model. As we experimentally illustrate in \cref{sec:ablation}, we utilize a Long-Short-Term-Memory (LSTM) model as our pose decoder since it has less memory requirement than transformer architectures and provides better auto-regressive capabilities. Finally, we map the output of the autoregressive model back to the pose space using an MLP layer. 

%% file: sec/5_experiments.tex
\section{Experiments}
\input{Tables/dataset_fitting}

\subsection{Dataset Evaluation}
Given that accurate 3D annotations are a requisite for the training of a potent SLP model, we quantitatively and qualitatively evaluated the performance of the pipeline introduced in \cref{sec:fitting} on the task of sign language reconstruction from videos. To assess the fitting quality, we apply the proposed pipeline to the SGNify mocap dataset \cite{SGNify} that contains ground truth annotations. 
\begin{figure}[!ht]
    \centering    \includegraphics[width=0.9\linewidth]{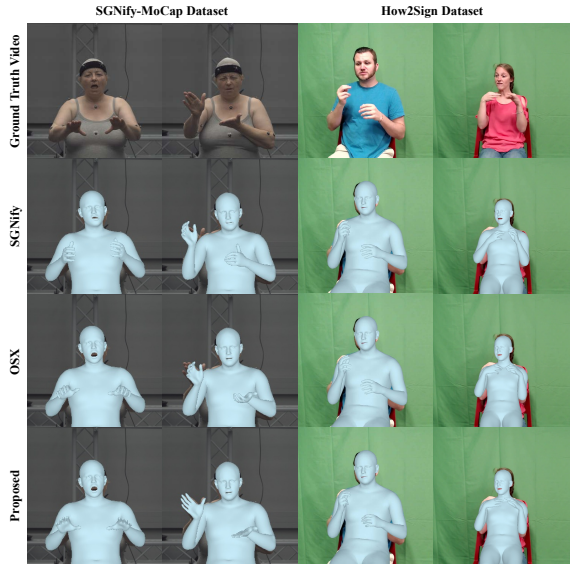}
    \captionof{figure}{\small{Qualitative comparison between the proposed and the baseline fitting frameworks on SGNify \cite{SGNify} and How2Sign \cite{duarte2021how2sign}.}
    \label{fig:fitting_comparison}}
\end{figure}
In \cref{tab:fitting_comparisons}, we report the reconstruction error of the proposed pipeline and compare it with OSX \cite{lin2023one} and the SGNify method, which is the current state-of-the-art method for 3D fitting from SL videos. It must be noted that unlike the SGNify method, the proposed fitting pipeline achieves a smaller reconstruction error despite not including any SL-driven losses. The powerful prior model that guides the fitting optimization leads our method to valid poses and articulation \cref{fig:fitting_comparison}.

\input{Tables/how2sign_evalutation}
\begin{figure*}[!ht]
    \centering
    \includegraphics[width=0.9\textwidth]{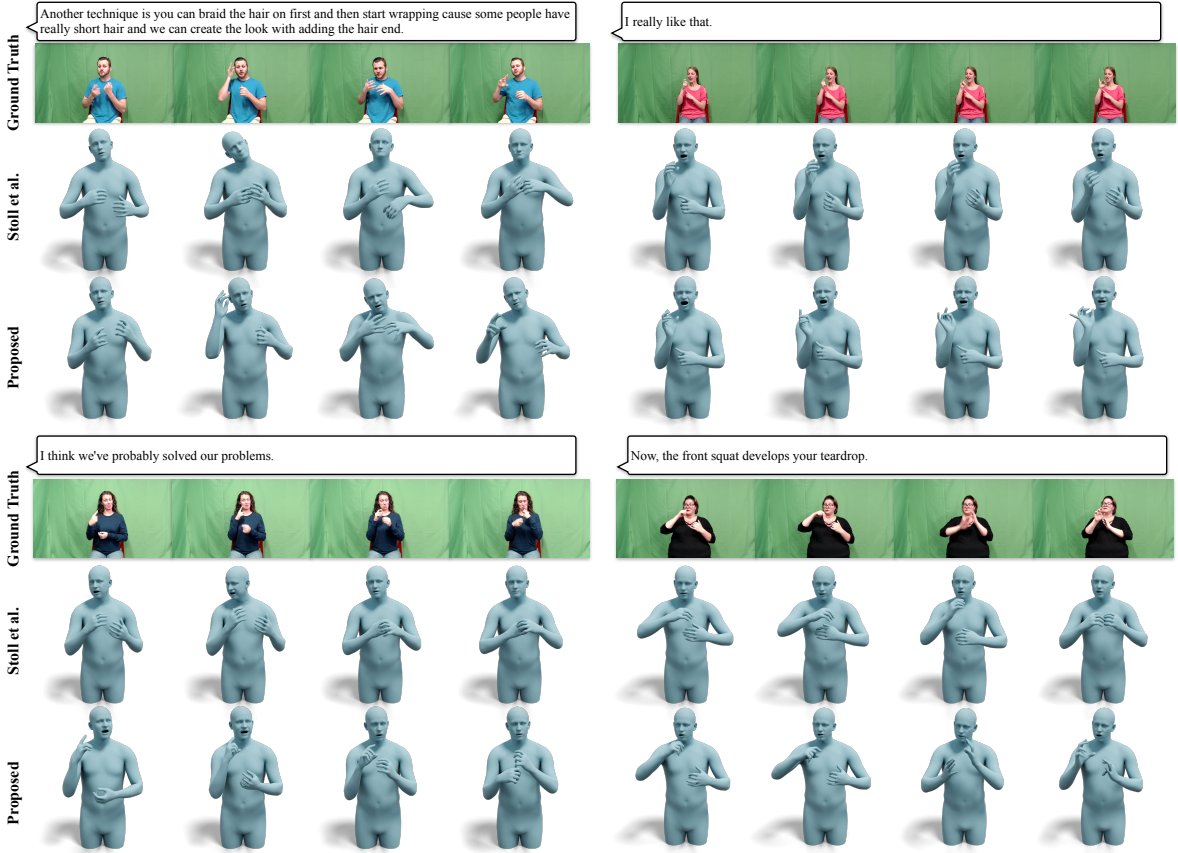}
    \captionof{figure}{\small{Qualitative comparison of generated signs conditioned on the text transcript between the proposed and Stoll \etal \cite{stoll2022there} methods. The ground truth video is given for reference. }
    \label{fig:main_figure}}
\end{figure*}
\subsection{Sign Language Production}
\noindent\textbf{Baselines.} To evaluate the performance of our method we selected the current state-of-the-art methods for text-driven sign language generation, \ie Saunders \etal \cite{saunders2020progressive}, Saunders \etal \cite{saunders2020adversarial}, Hwang \etal \cite{hwang2021non} and Stoll \etal \cite{stoll2022there}. Given that all methods have been trained on the German Sign Language RWTH-Phoenix dataset \cite{camgoz2018neural}, we retrained the models using the same training set-up as the proposed method.  

\noindent\textbf{Implementation Details.}
To train the diffusion model we followed the implementation details of \cite{ho2020denoising}. We implemented pose and expression embedding layers using a simple linear projection. Pose and expression encoders are composed of 4 stacked GNN and MLP layers, respectively, with an increasing number of channels. We employed the CLIP-ViT-L-14 model as our text encoder. The RNN decoder consists of 4 LSTM layers. We trained our model for 2K epochs using the Adam optimizer with a linearly decreasing learning rate from $10^{-3}$ to $10^{-6}$. 

\noindent\textbf{Evaluation Metrics.} 
We quantitatively evaluate the generation quality of the proposed and the baseline methods under a set of metrics. The first two were Mean Per Vertex Position Error (MPVPE) and Mean Per Joint Position Error (MPJPE).
Given that the motions generated from the proposed and baseline methods may not be correctly aligned with the ground truth annotations, we used Dynamic Time Warping (DTW) \cite{berndt1994using} to measure the similarity between the generated and the original sign sequences. To evaluate the quality of the generated poses we measured the Fréchet inception distance (FID) score between the generated and the ground truth poses. 
Finally, following \cite{saunders2020progressive}, we trained a Transformer-based back-translation network to map pose sequences back to text. For additional details, we refer the reader to the supplementary material. 

\begin{figure}[!h]
    \centering
\includegraphics[width=0.9\linewidth]{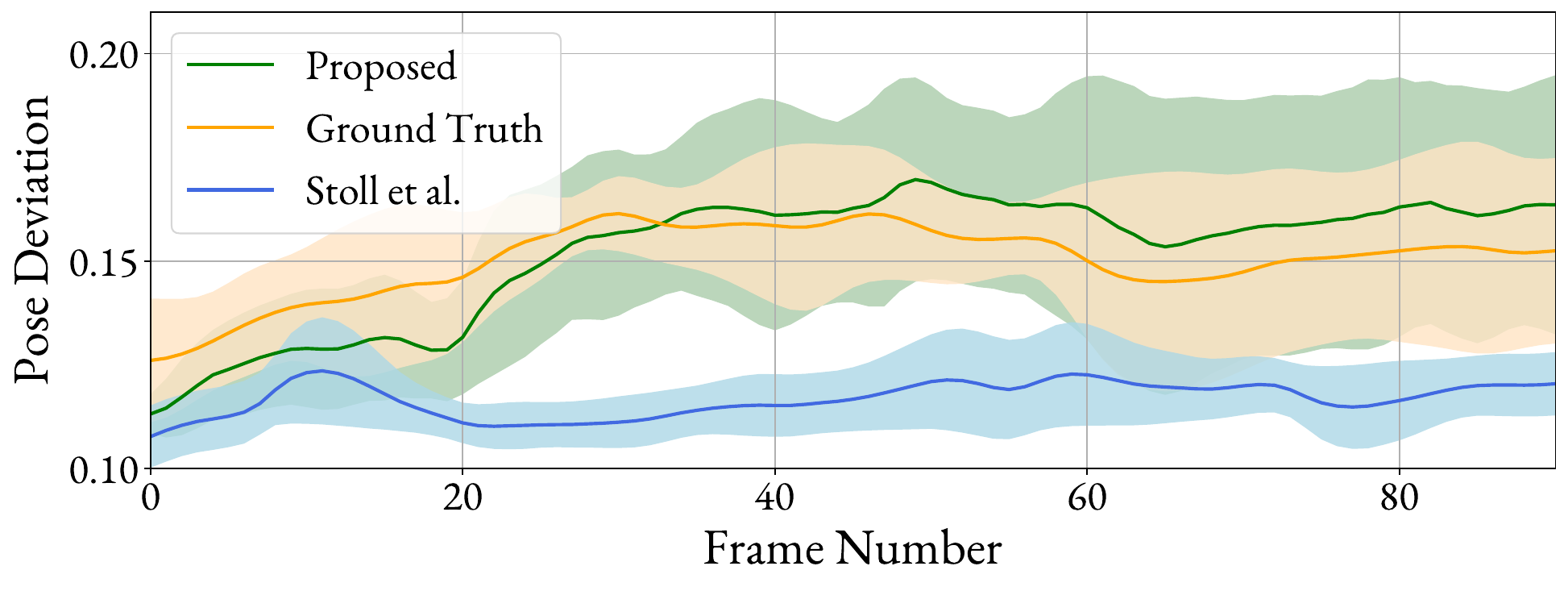}
    \vspace{-0.3cm}
    \captionof{figure}{\small{Mean and Standard Deviation of the absolute pose value across the sequence.}
    \label{fig:frame_diversity}}
\end{figure}

\noindent\textbf{Evaluation of Generated Signs.}
In the first three columns of \cref{tab:how2sign_eval} we quantitatively compare the proposed and the baseline methods on the test set of the curated dataset. The proposed method manages to outperform the baselines under all metrics, even by a large margin.  
Specifically, the generated signs not only demonstrate low reconstruction error across the entire upper body but also exhibit significant improvements on the hands region. Additionally, the proposed method is able to generate articulations that match the ground truth signs, which is translated to low FID scores. This can be also validated in \cref{fig:main_figure}, where the proposed method is able to generate signs with high-frequency articulations that match the ground truth videos. In contrast, current state-of-the-art SLP methods fail to model high-frequency articulations and can only generate small deviations around the canonical pose. This is quantified in \cref{fig:frame_diversity}, where we report the average per-frame pose deviations. The proposed method not only produces a larger variety of poses, compared to the small deviations of Stoll \etal \cite{stoll2022there},  but also follows the ground truth pose distribution. To enhance readability, we focus on Stoll \etal for qualitative comparisons, as the best performing prior work. 

\noindent\textbf{3D Pose Back-Translation.}
We additionally evaluated the overall quality of the generated pose sequences using back-translation, which measures how much information of the input sentences has been maintained in the model's output. In particular, we trained a Transformer-based architecture on our curated How2Sign dataset to learn a mapping from pose sequences back to the original text transcripts.
We then translated the generated 3D pose sequences of How2Sign back to spoken language using our back-translation network. To comply with the evaluations in \cite{saunders2020progressive,stoll2022there} we report BLEU n-grams from 1 to 4 and ROUGE scores. 
We repeated the above process for sequences generated by our model, as well as the baseline models \cite{saunders2020adversarial,saunders2020progressive,hwang2021non,stoll2022there} and summarize the results in the last column of \cref{tab:how2sign_eval}.
Our model produces the highest back translation scores across all metrics, with BLEU-4 being at the level of BLEU-2 of the second best performing method (Stoll \etal \cite{stoll2022there}).

\subsection{User-Study}
Although evaluation metrics can provide insights into a network's performance, the most critical benchmark lies in the perceptual evaluation from Hard of Hearing individuals. Notably, we further assess the realism of the generated signs by designing a user study where 15 ASL fluent subjects, with ages ranging from 29 to 62, evaluated the generated signs. We divided the perceptual study into two parts, to assess: (a) how aligned the generated signs are with respect to the text transcripts and (b) the fidelity and readability of the proposed generations. For the first part of the user study, we presented 15 different generated signs from both the proposed and baseline methods, alongside the ground truth video and its corresponding fitting. Participants were asked to assign a value between 1-10 rating the alignment of each method with the corresponding text transcript. 

\begin{figure}[!ht]
    \centering
\includegraphics[width=0.9\linewidth]{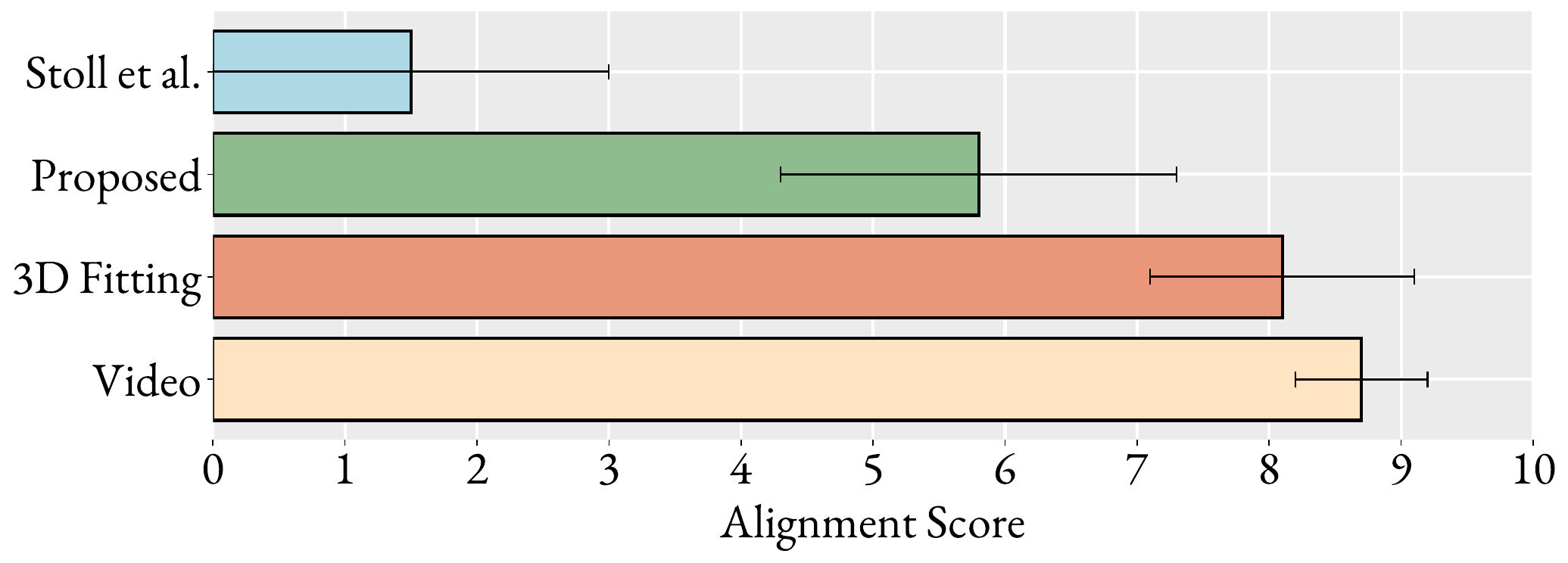}
    \vspace{-0.2cm}
    \captionof{figure}{\small{Human Evaluation of the alignment between the generated signs and the text transcript.}
    \label{fig:user_study}}
\end{figure}

To avoid potential biases between the methods, all videos were shown in a random order. 
In \cref{fig:user_study}, we report the results of the first part of the user-study. As expected, the ground truth videos achieve the best average score of 8.7 while the fittings achieve slightly less with 8.1,  which quantifies the high quality of the generated annotations. The proposed method achieves an average score of 5.8 whereas the method of Stoll \etal \cite{stoll2022there} fails to achieve reasonable results. 

The second part of the perceptual study aimed to assess the fidelity of the generated signs. Each participant was shown 15 rendered videos with signs generated by the proposed method and was asked to rank five candidate text translations, from most likely to least likely translation. Apart from the ground truth, the candidate translations included both similar to the ground truth sentences with slightly modified meanings, \ie by masking words, cropping sentences, or changing the word order, and also unrelated sentences from different topics and domains. Measuring the cumulative accuracy, the generated signs attained a top-1 accuracy of 40\% and a top-2 accuracy of 80\%, affirming the realism and fidelity of the generated signs.

\subsection{Ablation}
\label{sec:ablation}
The proposed method consists of four main components: the anatomically inspired pose encoder, the pose and expression embeddings, the autoregressive decoder, and the text encoding. In this section, we evaluate the contribution of each component to the final generations of the model. 

\noindent\textbf{Effect of Pose Encoder and Embedding Layers.} Initially, we ablate the novel pose encoder and we substitute it with an MLP layer (\textit{w/o GNN}), similar to the expression encoder. Under this setting, the generated motions are smoother and mainly deviate around the mean pose. Aligned with our hypothesis, updating poses in an anatomically inspired manner enables the network to learn higher frequencies, such as rare articulations, without an increase in the network's capacity. Additionally, as mentioned in \cref{sec:gnn}, in the traditional setting of motion diffusion models, during the denoising step, poses and expressions are sampled from a Gaussian distribution and are then decoded back to their original space using a permutation equivariant network, such as an MLP. Such permutation equivariance, treats poses under a uniform setting that limits the generative power of the network. As shown in \cref{tab:ablation}, without pose and expression embedding layer (\textit{w/o Pose and Expression Embedding}), the model fails to produce any reasonable sign, resulting in a performance drop under all metrics. 
\input{Tables/ablation}

\noindent\textbf{Effect of LSTM Encoder.}
A core part of the proposed method is the autoregressive LSTM decoder. We initially evaluate its contribution compared to a simple frame positional encoding to transform the network to a frame-conditioned generative model, without having any temporal module (\textit{w/o LSTM}). As expected, this results in poor DTW performance and generations that present increased jittering and lack of temporal coherence. Furthermore, we substitute the LSTM layer with a Transformer encoder layer (\textit{w. Transformer}). Interestingly, the LSTM layer achieves similar performance to the Transformer layer while having 75\% fewer parameters (1M vs. 4.5M).

\noindent\textbf{Effect of Text Encoding.}
The text encoding module has a pivotal effect on motion generation. Firstly, we substitute the CLIP encoder with a word-level embedding layer that is trained with the rest of the method in an end-to-end fashion. We set the embedded size to 256 although we did not observe significant differences in performance. Following this, we utilized the pretrained DistilBERT \cite{sanh2019distilbert}, whose parameters remain frozen throughout training. As depicted in \cref{tab:ablation}, training word embeddings from scratch strongly affects the generalization of the network, leading to large MPVPE and MPJPE metrics. In contrast, DistilBERT achieves better performance than learnable word embeddings, although it does not outperform CLIP embeddings. This is aligned with our assumption that sentence embeddings could provide better insights regarding the meaning of a sentence compared to word-level embeddings. Especially in the task of SLP, where there is not an explicit one-to-one mapping between words and poses, sentence level embeddings provide a more powerful text encoding solution. 

\noindent\textbf{Comparison with Human Motion Diffusion Models.} Finally, we compare our model with state-of-the-art methods on human motion modeling \cite{tevet2022human,chen2023executing}, which can be considered as deviations from the proposed framework. Unlike our anatomically inspired approach, both models rely on linear layers to handle pose motions, constraining their capacity to encode intricate hand movements with high-frequency details. In particular, although both methods can achieve smooth body motions, they fail to produce accurate hand articulations that match the ground truth distribution, which can be validated from the reconstruction errors (MPVPE, MPVJE), along with the FID measure.

%% file: Tables/dataset_fitting.tex
\begin{table}[t]
    \centering
    \caption{Mean per vertex error (mm) of the proposed and the baseline methods on the SGNify mocap dataset \cite{SGNify}.}
    \resizebox{0.75\linewidth}{!}{
    \begin{tabular}{l|ccc}
    \toprule
    \textbf{Method} & \textbf{Body} & \textbf{Left Hand} & \textbf{Right Hand} 
    \\
    \midrule 
        FrankMoCap\cite{rong2021frankmocap} & 78.07 & 20.47 & 19.62  \\
        PIXIE\cite{PIXIE}  & 60.11 & 25.02 & 22.42 \\
        PyMAF-X~\cite{pymafx2023}  & 68.61 & 21.46 & 19.19 \\
        SMPLify-X \cite{SMPL-X} & 56.07 & 22.23 & 18.83 \\
        SGNify \cite{SGNify} & {55.63} & {19.22} & {17.50} \\
        OSX \cite{lin2023one} & {47.32} & {18.34} & {18.12} \\
        \textbf{Proposed}  & \textbf{46.42} &  \textbf{16.17} &  \textbf{15.23} \\
    \bottomrule
    \end{tabular}}
    \label{tab:fitting_comparisons}
\end{table}

%% file: Tables/how2sign_evalutation.tex
\begin{table*}[t]
    \centering
    \caption{Quantitative evaluation of the proposed and the baseline methods on the How2Sign test dataset. }
    \setlength{\tabcolsep}{2pt}
    \resizebox{0.9\linewidth}{!}{
    \begin{tabular}{l|cccc|cccc|cccc|ccccc}
    \midrule
       & \multicolumn{4}{c|}{Body} & \multicolumn{4}{c|}{Left Hand} & \multicolumn{4}{c|}{Right Hand} & \multicolumn{5}{c}{Back-Translation}
    \\
    \toprule 
     \textbf{Method} & \textbf{MPVPE} $\downarrow$ & \textbf{MPJPE} $\downarrow$ & \textbf{FID} $\downarrow$ & \textbf{DTW} $\downarrow$ & \textbf{MPVPE} & \textbf{MPJPE} & \textbf{FID} & \textbf{DTW} & \textbf{MPVPE} & \textbf{MPJPE} & \textbf{FID} & \textbf{DTW} & \textbf{BLEU-4} $\uparrow$ & \textbf{BLEU-3} $\uparrow$ & \textbf{BLEU-2} $\uparrow$ & \textbf{BLEU-1} $\uparrow$ & \textbf{ROUGE} $\uparrow$
    \\
    \midrule 
        Saunders \etal \cite{saunders2020progressive}  & 
        67.21 & 70.06 & 4.71 & 14.15 & 
        73.49 & 74.13 & 0.68 & 11.21 & 
        75.57 & 77.47 & 0.75 & 11.93 & 
        2.75  & 5.87  & 8.21 & 13.82 & 29.87 \\
        
        Saunders \etal \cite{saunders2020adversarial} & 
        63.19 & 65.25 & 3.98 & 13.78 & 
        71.43 & 72.39 & 0.59 & 11.02 & 
        68.54 & 70.14 & 0.51 & 11.32 &
        6.21  & 8.98  & 12.01 & 18.22 & 32.33 \\

        Hwang \etal \cite{hwang2021non} & 
        62.74 & 63.25 & 4.45 & 13.94 &
        78.95 & 70.34 & 0.63 & 11.33 &
        68.65 & 69.59 & 0.60 & 12.26 &
        5.75  & 8.21  & 11.62 & 17.55 & 31.98 \\ 
        
        Stoll \etal \cite{stoll2022there} &
        55.02 & 60.32 & 4.96 & 13.99 & 
        68.48 & 69.45 & 0.56 & 11.59 & 
        60.18 & 62.73 & 0.64 & 12.29 &
        7.51  & 10.72 & 13.92 & 19.56& 33.17 \\
        
        \textbf{Proposed} &
        \textbf{31.47} & \textbf{35.87} & \textbf{1.56} & \textbf{7.83} & 
        \textbf{36.24} & \textbf{38.82} & \textbf{0.24} & \textbf{6.74} & 
        \textbf{39.68} & \textbf{40.56} & \textbf{0.36} & \textbf{7.91} & \textbf{13.12} & \textbf{18.25} & \textbf{25.44} & \textbf{41.31} & \textbf{47.55}\\
    \bottomrule
    \end{tabular}
    }
    \label{tab:how2sign_eval}
\end{table*}

%% file: Tables/ablation.tex
\begin{table}[t]
    \centering
    \caption{Evaluation of individual components in the proposed method. Every row refers to a different ablated module. We include the performance of our method for reference.}
    \resizebox{\linewidth}{!}{
    \begin{tabular}{l|cccc}
    \toprule
    \textbf{Method} & \textbf{MPVPE} & \textbf{MPJPE} & \textbf{FID} & \textbf{DTW}
    \\
    \midrule 
        w/o {GNN}        & 37.51 & 38.23 & 2.85  & 9.19\\
        w/o Pose and Expression Embedding 
                         & 66.56 & 68.34 & 6.65  & 11.98\\
        \hline
        w/o LSTM         & 36.17 & 39.46 & 2.12 & 10.82 \\
        w. Transformer   & 32.73 & \textbf{35.41} & 1.58 & 8.17  \\
        \hline
        w/o CLIP         & 69.12 & 71.42 & 5.36  & 12.84\\
        w. BERT          & 45.32 & 47.11 & 2.23  &  9.21 \\
        \hline
        Tevet \etal \cite{tevet2022human}  & 36.11 & 38.23 & 2.45  & 8.92\\ 
        Chen \etal \cite{chen2023executing} &  35.23 & 37.12 & 2.15  & 8.26\\
        \hline
        \textbf{Proposed} & \textbf{31.47} & {35.87} & \textbf{1.56} & \textbf{7.83}\\
    \bottomrule
    \end{tabular}}
    \label{tab:ablation}
\end{table}

%% file: sec/6_conclusion.tex
\section{Conclusion}
Neural 3D sign language production is an important challenge that aims to aid the Deaf and Hard of Hearing community and can effectively increase their inclusion in any social environment. In this work, we made a step towards high fidelity 3D SLP, by deriving a large-scale 3D dataset to train a text conditioned diffusion-based model. The release of additional relevant databases will enable the training of even more robust architectures. We initially introduce a precise 3D sign language reconstruction pipeline that outperforms previous SL reconstruction methods. Then, we train a motion generative model using an autoregressive diffusion model. The core of our method is founded on a novel, anatomically inspired, graph neural network that learns the pose distribution and enables highly detailed articulations. Importantly, leveraging the powerful CLIP text embeddings, the proposed model can generalize to out-of-distribution samples. Extensive experiments on sign language generation tasks, including a perceptual study with ASL fluent subjects, demonstrate the superiority of the proposed method compared to the previous approaches. 

\noindent\textbf{Acknowledgements.} S. Zafeiriou was supported by EPSRC Project DEFORM (EP/S010203/1) and GNOMON (EP/X011364). R.A. Potamias was supported by EPSRC Project GNOMON (EP/X011364).

%% file: main.bbl
\begin{thebibliography}{64}
\providecommand{\natexlab}[1]{#1}
\providecommand{\url}[1]{\texttt{#1}}
\expandafter\ifx\csname urlstyle\endcsname\relax
  \providecommand{\doi}[1]{doi: #1}\else
  \providecommand{\doi}{doi: \begingroup \urlstyle{rm}\Url}\fi

\bibitem[Agris and Kraiss(2010)]{agris2010signum}
Ulrich~von Agris and Karl-Friedrich Kraiss.
\newblock Signum database: Video corpus for signer-independent continuous sign language recognition.
\newblock In \emph{sign-lang@ LREC 2010}, pages 243--246. European Language Resources Association (ELRA), 2010.

\bibitem[Albanie et~al.(2020)Albanie, Varol, Momeni, Afouras, Chung, Fox, and Zisserman]{albanie2020bsl}
Samuel Albanie, G{\"u}l Varol, Liliane Momeni, Triantafyllos Afouras, Joon~Son Chung, Neil Fox, and Andrew Zisserman.
\newblock Bsl-1k: Scaling up co-articulated sign language recognition using mouthing cues.
\newblock In \emph{Computer Vision--ECCV 2020: 16th European Conference, Glasgow, UK, August 23--28, 2020, Proceedings, Part XI 16}, pages 35--53. Springer, 2020.

\bibitem[Antonakos et~al.(2015)Antonakos, Roussos, and Zafeiriou]{antonakos2015survey}
Epameinondas Antonakos, Anastasios Roussos, and Stefanos Zafeiriou.
\newblock A survey on mouth modeling and analysis for sign language recognition.
\newblock In \emph{2015 11th IEEE International Conference and Workshops on Automatic Face and Gesture Recognition (FG)}, pages 1--7. IEEE, 2015.

\bibitem[Athitsos et~al.(2008)Athitsos, Neidle, Sclaroff, Nash, Stefan, Yuan, and Thangali]{athitsos2008american}
Vassilis Athitsos, Carol Neidle, Stan Sclaroff, Joan Nash, Alexandra Stefan, Quan Yuan, and Ashwin Thangali.
\newblock The american sign language lexicon video dataset.
\newblock In \emph{2008 IEEE Computer Society Conference on Computer Vision and Pattern Recognition Workshops}, pages 1--8. IEEE, 2008.

\bibitem[Berndt and Clifford(1994)]{berndt1994using}
Donald~J Berndt and James Clifford.
\newblock Using dynamic time warping to find patterns in time series.
\newblock In \emph{Proceedings of the 3rd international conference on knowledge discovery and data mining}, pages 359--370, 1994.

\bibitem[Braem and Sutton-Spence(2001)]{braem2001hands}
P~Boyes Braem and RL Sutton-Spence.
\newblock \emph{The Hands Are The Head of The Mouth. The Mouth as Articulator in Sign Languages}.
\newblock Hamburg: Signum Press, 2001.

\bibitem[Braffort et~al.(2010)Braffort, Bolot, Ch{\'e}telat-Pel{\'e}, Choisier, Delorme, Filhol, Segouat, Verrecchia, Badin, and Devos]{braffort2010sign}
Annelies Braffort, Laurence Bolot, Emilie Ch{\'e}telat-Pel{\'e}, Annick Choisier, Maxime Delorme, Michael Filhol, J{\'e}r{\'e}mie Segouat, Cyril Verrecchia, Flora Badin, and Nad{\`e}ge Devos.
\newblock Sign language corpora for analysis, processing and evaluation.
\newblock In \emph{LREC}, 2010.

\bibitem[Bragg et~al.(2019)Bragg, Koller, Bellard, Berke, Boudreault, Braffort, Caselli, Huenerfauth, Kacorri, Verhoef, et~al.]{bragg2019sign}
Danielle Bragg, Oscar Koller, Mary Bellard, Larwan Berke, Patrick Boudreault, Annelies Braffort, Naomi Caselli, Matt Huenerfauth, Hernisa Kacorri, Tessa Verhoef, et~al.
\newblock Sign language recognition, generation, and translation: An interdisciplinary perspective.
\newblock In \emph{Proceedings of the 21st International ACM SIGACCESS Conference on Computers and Accessibility}, pages 16--31, 2019.

\bibitem[Camgoz et~al.(2018)Camgoz, Hadfield, Koller, Ney, and Bowden]{camgoz2018neural}
Necati~Cihan Camgoz, Simon Hadfield, Oscar Koller, Hermann Ney, and Richard Bowden.
\newblock Neural sign language translation.
\newblock In \emph{Proceedings of the IEEE conference on computer vision and pattern recognition}, pages 7784--7793, 2018.

\bibitem[Camgoz et~al.(2020)Camgoz, Koller, Hadfield, and Bowden]{camgoz2020sign}
Necati~Cihan Camgoz, Oscar Koller, Simon Hadfield, and Richard Bowden.
\newblock Sign language transformers: Joint end-to-end sign language recognition and translation.
\newblock In \emph{Proceedings of the IEEE/CVF conference on computer vision and pattern recognition}, pages 10023--10033, 2020.

\bibitem[{Cao} et~al.(2019){Cao}, {Hidalgo Martinez}, {Simon}, {Wei}, and {Sheikh}]{OPENPOSE}
Z. {Cao}, G. {Hidalgo Martinez}, T. {Simon}, S. {Wei}, and Y.~A. {Sheikh}.
\newblock Openpose: Realtime multi-person 2d pose estimation using part affinity fields.
\newblock \emph{IEEE Transactions on Pattern Analysis and Machine Intelligence}, 2019.

\bibitem[Chai et~al.(2014)Chai, Wang, and Chen]{chai2014devisign}
Xiujuan Chai, Hanjie Wang, and Xilin Chen.
\newblock The devisign large vocabulary of chinese sign language database and baseline evaluations.
\newblock In \emph{Technical report VIPL-TR-14-SLR-001. Key Lab of Intelligent Information Processing of Chinese Academy of Sciences (CAS)}. Institute of Computing Technology, 2014.

\bibitem[Chen et~al.(2023)Chen, Jiang, Liu, Huang, Fu, Chen, and Yu]{chen2023executing}
Xin Chen, Biao Jiang, Wen Liu, Zilong Huang, Bin Fu, Tao Chen, and Gang Yu.
\newblock Executing your commands via motion diffusion in latent space.
\newblock In \emph{Proceedings of the IEEE/CVF Conference on Computer Vision and Pattern Recognition}, pages 18000--18010, 2023.

\bibitem[Chen et~al.(2022{\natexlab{a}})Chen, Wei, Sun, Wu, and Lin]{chen2022simple}
Yutong Chen, Fangyun Wei, Xiao Sun, Zhirong Wu, and Stephen Lin.
\newblock A simple multi-modality transfer learning baseline for sign language translation.
\newblock In \emph{Proceedings of the IEEE/CVF Conference on Computer Vision and Pattern Recognition}, pages 5120--5130, 2022{\natexlab{a}}.

\bibitem[Chen et~al.(2022{\natexlab{b}})Chen, Zuo, Wei, Wu, Liu, and Mak]{chen2022two}
Yutong Chen, Ronglai Zuo, Fangyun Wei, Yu Wu, Shujie Liu, and Brian Mak.
\newblock Two-stream network for sign language recognition and translation.
\newblock \emph{Advances in Neural Information Processing Systems}, 35:\penalty0 17043--17056, 2022{\natexlab{b}}.

\bibitem[Cox et~al.(2002)Cox, Lincoln, Tryggvason, Nakisa, Wells, Tutt, and Abbott]{cox2002tessa}
Stephen Cox, Michael Lincoln, Judy Tryggvason, Melanie Nakisa, Mark Wells, Marcus Tutt, and Sanja Abbott.
\newblock Tessa, a system to aid communication with deaf people.
\newblock In \emph{Proceedings of the fifth international ACM conference on Assistive technologies}, pages 205--212, 2002.

\bibitem[Davydov et~al.(2022)Davydov, Remizova, Constantin, Honari, Salzmann, and Fua]{davydov2022adversarial}
Andrey Davydov, Anastasia Remizova, Victor Constantin, Sina Honari, Mathieu Salzmann, and Pascal Fua.
\newblock Adversarial parametric pose prior.
\newblock In \emph{Proceedings of the IEEE/CVF Conference on Computer Vision and Pattern Recognition}, pages 10997--11005, 2022.

\bibitem[Dreuw et~al.(2007)Dreuw, Rybach, Deselaers, Zahedi, and Ney]{dreuw2007speech}
Philippe Dreuw, David Rybach, Thomas Deselaers, Morteza Zahedi, and Hermann Ney.
\newblock Speech recognition techniques for a sign language recognition system.
\newblock \emph{hand}, 60:\penalty0 80, 2007.

\bibitem[Duarte et~al.(2021)Duarte, Palaskar, Ventura, Ghadiyaram, DeHaan, Metze, Torres, and Giro-i Nieto]{duarte2021how2sign}
Amanda Duarte, Shruti Palaskar, Lucas Ventura, Deepti Ghadiyaram, Kenneth DeHaan, Florian Metze, Jordi Torres, and Xavier Giro-i Nieto.
\newblock How2sign: a large-scale multimodal dataset for continuous american sign language.
\newblock In \emph{Proceedings of the IEEE/CVF conference on computer vision and pattern recognition}, pages 2735--2744, 2021.

\bibitem[Efthimiou et~al.(2010)Efthimiou, Fotinea, Hanke, Glauert, Bowden, Braffort, Collet, Maragos, and Goudenove]{efthimiou2010dicta}
Eleni Efthimiou, Stavroula-Evita Fotinea, Thomas Hanke, John Glauert, Richard Bowden, Annelies Braffort, Christophe Collet, Petros Maragos, and Fran{\c{c}}ois Goudenove.
\newblock Dicta-sign: sign language recognition, generation and modelling with application in deaf communication.
\newblock In \emph{sign-lang@ LREC 2010}, pages 80--83. European Language Resources Association (ELRA), 2010.

\bibitem[Efthimiou et~al.(2012)Efthimiou, Fotinea, Hanke, Glauert, Bowden, Braffort, Collet, Maragos, and Lefebvre-Albaret]{efthimiou2012dicta}
Eleni Efthimiou, Stavroula-Evita Fotinea, Thomas Hanke, John Glauert, Richard Bowden, Annelies Braffort, Christophe Collet, Petros Maragos, and Fran{\c{c}}ois Lefebvre-Albaret.
\newblock The dicta-sign wiki: Enabling web communication for the deaf.
\newblock In \emph{Computers Helping People with Special Needs: 13th International Conference, ICCHP 2012, Linz, Austria, July 11-13, 2012, Proceedings, Part II 13}, pages 205--212. Springer, 2012.

\bibitem[Fan et~al.(2023)Fan, Taheri, Tzionas, Kocabas, Kaufmann, Black, and Hilliges]{fan2023arctic}
Zicong Fan, Omid Taheri, Dimitrios Tzionas, Muhammed Kocabas, Manuel Kaufmann, Michael~J. Black, and Otmar Hilliges.
\newblock {ARCTIC}: A dataset for dexterous bimanual hand-object manipulation.
\newblock In \emph{Proceedings IEEE Conference on Computer Vision and Pattern Recognition (CVPR)}, 2023.

\bibitem[Feng et~al.(2021)Feng, Choutas, Bolkart, Tzionas, and Black]{PIXIE}
Yao Feng, Vasileios Choutas, Timo Bolkart, Dimitrios Tzionas, and Michael Black.
\newblock Collaborative regression of expressive bodies using moderation.
\newblock In \emph{International Conference on 3D Vision (3DV)}, pages 792--804, 2021.

\bibitem[Forster et~al.(2014)Forster, Schmidt, Koller, Bellgardt, and Ney]{forster2014extensions}
Jens Forster, Christoph Schmidt, Oscar Koller, Martin Bellgardt, and Hermann Ney.
\newblock Extensions of the sign language recognition and translation corpus rwth-phoenix-weather.
\newblock In \emph{LREC}, pages 1911--1916, 2014.

\bibitem[Forte et~al.(2023)Forte, Kulits, Huang, Choutas, Tzionas, Kuchenbecker, and Black]{SGNify}
Maria-Paola Forte, Peter Kulits, Chun-Hao~Paul Huang, Vasileios Choutas, Dimitrios Tzionas, Katherine~J. Kuchenbecker, and Michael~J. Black.
\newblock Reconstructing signing avatars from video using linguistic priors.
\newblock In \emph{IEEE/CVF Conf.~on Computer Vision and Pattern Recognition (CVPR)}, pages 12791--12801, 2023.

\bibitem[Grathwohl et~al.(2018)Grathwohl, Chen, Bettencourt, Sutskever, and Duvenaud]{grathwohl2018ffjord}
Will Grathwohl, Ricky~TQ Chen, Jesse Bettencourt, Ilya Sutskever, and David Duvenaud.
\newblock Ffjord: Free-form continuous dynamics for scalable reversible generative models.
\newblock \emph{arXiv preprint arXiv:1810.01367}, 2018.

\bibitem[Guo et~al.(2022)Guo, Zou, Zuo, Wang, Ji, Li, and Cheng]{guo2022generating}
Chuan Guo, Shihao Zou, Xinxin Zuo, Sen Wang, Wei Ji, Xingyu Li, and Li Cheng.
\newblock Generating diverse and natural 3d human motions from text.
\newblock In \emph{Proceedings of the IEEE/CVF Conference on Computer Vision and Pattern Recognition}, pages 5152--5161, 2022.

\bibitem[Hao et~al.(2021)Hao, Min, and Chen]{hao2021self}
Aiming Hao, Yuecong Min, and Xilin Chen.
\newblock Self-mutual distillation learning for continuous sign language recognition.
\newblock In \emph{Proceedings of the IEEE/CVF International Conference on Computer Vision}, pages 11303--11312, 2021.

\bibitem[Ho et~al.(2020)Ho, Jain, and Abbeel]{ho2020denoising}
Jonathan Ho, Ajay Jain, and Pieter Abbeel.
\newblock Denoising diffusion probabilistic models.
\newblock \emph{Advances in neural information processing systems}, 33:\penalty0 6840--6851, 2020.

\bibitem[Hwang et~al.(2021)Hwang, Kim, and Park]{hwang2021non}
Eui~Jun Hwang, Jung-Ho Kim, and Jong~C. Park.
\newblock Non-autoregressive sign language production with gaussian space.
\newblock In \emph{The 32nd British Machine Vision Conference (BMVC 21)}. British Machine Vision Conference (BMVC), 2021.

\bibitem[Joze and Koller(2018)]{joze2018ms}
Hamid Reza~Vaezi Joze and Oscar Koller.
\newblock Ms-asl: A large-scale data set and benchmark for understanding american sign language.
\newblock \emph{arXiv preprint arXiv:1812.01053}, 2018.

\bibitem[Kapoor et~al.(2021)Kapoor, Mukhopadhyay, Hegde, Namboodiri, and Jawahar]{kapoor2021towards}
Parul Kapoor, Rudrabha Mukhopadhyay, Sindhu~B Hegde, Vinay Namboodiri, and CV Jawahar.
\newblock Towards automatic speech to sign language generation.
\newblock \emph{arXiv preprint arXiv:2106.12790}, 2021.

\bibitem[Li et~al.(2020{\natexlab{a}})Li, Rodriguez, Yu, and Li]{li2020word}
Dongxu Li, Cristian Rodriguez, Xin Yu, and Hongdong Li.
\newblock Word-level deep sign language recognition from video: A new large-scale dataset and methods comparison.
\newblock In \emph{Proceedings of the IEEE/CVF winter conference on applications of computer vision}, pages 1459--1469, 2020{\natexlab{a}}.

\bibitem[Li et~al.(2020{\natexlab{b}})Li, Xu, Yu, Zhang, Swift, Suominen, and Li]{li2020tspnet}
Dongxu Li, Chenchen Xu, Xin Yu, Kaihao Zhang, Benjamin Swift, Hanna Suominen, and Hongdong Li.
\newblock Tspnet: Hierarchical feature learning via temporal semantic pyramid for sign language translation.
\newblock \emph{Advances in Neural Information Processing Systems}, 33:\penalty0 12034--12045, 2020{\natexlab{b}}.

\bibitem[Liang et~al.(2023)Liang, Li, and Chai]{SLT_survey}
Zeyu Liang, Huailing Li, and Jianping Chai.
\newblock Sign language translation: A survey of approaches and techniques.
\newblock \emph{Electronics}, 12\penalty0 (12):\penalty0 2678, 2023.

\bibitem[Lin et~al.(2023)Lin, Zeng, Wang, Zhang, and Li]{lin2023one}
Jing Lin, Ailing Zeng, Haoqian Wang, Lei Zhang, and Yu Li.
\newblock One-stage 3d whole-body mesh recovery with component aware transformer.
\newblock In \emph{Proceedings of the IEEE/CVF Conference on Computer Vision and Pattern Recognition}, pages 21159--21168, 2023.

\bibitem[Lugaresi et~al.(2019)Lugaresi, Tang, Nash, McClanahan, Uboweja, Hays, Zhang, Chang, Yong, Lee, et~al.]{lugaresi2019mediapipe}
Camillo Lugaresi, Jiuqiang Tang, Hadon Nash, Chris McClanahan, Esha Uboweja, Michael Hays, Fan Zhang, Chuo-Ling Chang, Ming~Guang Yong, Juhyun Lee, et~al.
\newblock Mediapipe: A framework for building perception pipelines.
\newblock \emph{arXiv preprint arXiv:1906.08172}, 2019.

\bibitem[Mahmood et~al.(2019)Mahmood, Ghorbani, F.~Troje, Pons-Moll, and Black]{AMASS}
Naureen Mahmood, Nima Ghorbani, Nikolaus F.~Troje, Gerard Pons-Moll, and Michael~J. Black.
\newblock Amass: Archive of motion capture as surface shapes.
\newblock In \emph{The IEEE International Conference on Computer Vision (ICCV)}, 2019.

\bibitem[McDonald et~al.(2016)McDonald, Wolfe, Schnepp, Hochgesang, Jamrozik, Stumbo, Berke, Bialek, and Thomas]{mcdonald2016automated}
John McDonald, Rosalee Wolfe, Jerry Schnepp, Julie Hochgesang, Diana~Gorman Jamrozik, Marie Stumbo, Larwan Berke, Melissa Bialek, and Farah Thomas.
\newblock An automated technique for real-time production of lifelike animations of american sign language.
\newblock \emph{Universal Access in the Information Society}, 15:\penalty0 551--566, 2016.

\bibitem[Naert et~al.(2020)Naert, Larboulette, and Gibet]{naert2020survey}
Lucie Naert, Caroline Larboulette, and Sylvie Gibet.
\newblock A survey on the animation of signing avatars: From sign representation to utterance synthesis.
\newblock \emph{Computers \& Graphics}, 92:\penalty0 76--98, 2020.

\bibitem[Pavlakos et~al.(2019)Pavlakos, Choutas, Ghorbani, Bolkart, Osman, Tzionas, and Black]{SMPL-X}
Georgios Pavlakos, Vasileios Choutas, Nima Ghorbani, Timo Bolkart, Ahmed~AA Osman, Dimitrios Tzionas, and Michael~J Black.
\newblock Expressive body capture: 3d hands, face, and body from a single image.
\newblock In \emph{Proceedings of the IEEE/CVF conference on computer vision and pattern recognition}, pages 10975--10985, 2019.

\bibitem[Potamias et~al.(2023)Potamias, Ploumpis, Moschoglou, Triantafyllou, and Zafeiriou]{potamias2023handy}
Rolandos~Alexandros Potamias, Stylianos Ploumpis, Stylianos Moschoglou, Vasileios Triantafyllou, and Stefanos Zafeiriou.
\newblock Handy: Towards a high fidelity 3d hand shape and appearance model.
\newblock In \emph{Proceedings of the IEEE/CVF Conference on Computer Vision and Pattern Recognition}, pages 4670--4680, 2023.

\bibitem[Radford et~al.(2021)Radford, Kim, Hallacy, Ramesh, Goh, Agarwal, Sastry, Askell, Mishkin, Clark, et~al.]{CLIP}
Alec Radford, Jong~Wook Kim, Chris Hallacy, Aditya Ramesh, Gabriel Goh, Sandhini Agarwal, Girish Sastry, Amanda Askell, Pamela Mishkin, Jack Clark, et~al.
\newblock Learning transferable visual models from natural language supervision.
\newblock In \emph{International conference on machine learning}, pages 8748--8763. PMLR, 2021.

\bibitem[Romero et~al.(2022)Romero, Tzionas, and Black]{romero2022embodied}
Javier Romero, Dimitrios Tzionas, and Michael~J Black.
\newblock Embodied hands: Modeling and capturing hands and bodies together.
\newblock \emph{arXiv preprint arXiv:2201.02610}, 2022.

\bibitem[Rong et~al.(2021)Rong, Shiratori, and Joo]{rong2021frankmocap}
Yu Rong, Takaaki Shiratori, and Hanbyul Joo.
\newblock Frankmocap: A monocular 3d whole-body pose estimation system via regression and integration.
\newblock In \emph{IEEE International Conference on Computer Vision Workshops}, 2021.

\bibitem[Sanabria et~al.(2018)Sanabria, Caglayan, Palaskar, Elliott, Barrault, Specia, and Metze]{sanabria2018how2}
Ramon Sanabria, Ozan Caglayan, Shruti Palaskar, Desmond Elliott, Lo{\"\i}c Barrault, Lucia Specia, and Florian Metze.
\newblock How2: a large-scale dataset for multimodal language understanding.
\newblock \emph{arXiv preprint arXiv:1811.00347}, 2018.

\bibitem[Sanh et~al.(2019)Sanh, Debut, Chaumond, and Wolf]{sanh2019distilbert}
Victor Sanh, Lysandre Debut, Julien Chaumond, and Thomas Wolf.
\newblock Distilbert, a distilled version of bert: smaller, faster, cheaper and lighter.
\newblock \emph{arXiv preprint arXiv:1910.01108}, 2019.

\bibitem[Saunders et~al.()Saunders, Camg{\"o}z, and Bowden]{saunders2020adversarial}
Ben Saunders, Necati~Cihan Camg{\"o}z, and Richard Bowden.
\newblock Adversarial training for multi-channel sign language production.
\newblock In \emph{The 31st British Machine Vision Virtual Conference}. British Machine Vision Association.

\bibitem[Saunders et~al.(2020)Saunders, Camgoz, and Bowden]{saunders2020progressive}
Ben Saunders, Necati~Cihan Camgoz, and Richard Bowden.
\newblock Progressive transformers for end-to-end sign language production.
\newblock In \emph{Computer Vision--ECCV 2020: 16th European Conference, Glasgow, UK, August 23--28, 2020, Proceedings, Part XI 16}, pages 687--705. Springer, 2020.

\bibitem[Saunders et~al.(2021)Saunders, Camgoz, and Bowden]{saunders2021mixed}
Ben Saunders, Necati~Cihan Camgoz, and Richard Bowden.
\newblock Mixed signals: Sign language production via a mixture of motion primitives.
\newblock In \emph{Proceedings of the IEEE/CVF International Conference on Computer Vision}, pages 1919--1929, 2021.

\bibitem[Saunders et~al.(2022)Saunders, Camgoz, and Bowden]{Saunders_2022_CVPR}
Ben Saunders, Necati~Cihan Camgoz, and Richard Bowden.
\newblock Signing at scale: Learning to co-articulate signs for large-scale photo-realistic sign language production.
\newblock In \emph{Proceedings of the IEEE/CVF Conference on Computer Vision and Pattern Recognition (CVPR)}, pages 5141--5151, 2022.

\bibitem[Schembri et~al.(2013)Schembri, Fenlon, Rentelis, Reynolds, and Cormier]{schembri2013building}
Adam Schembri, Jordan Fenlon, Ramas Rentelis, Sally Reynolds, and Kearsy Cormier.
\newblock Building the british sign language corpus.
\newblock 2013.

\bibitem[Stoll et~al.(2018)Stoll, Camg{\"o}z, Hadfield, and Bowden]{stoll2018sign}
Stephanie Stoll, Necati~Cihan Camg{\"o}z, Simon Hadfield, and Richard Bowden.
\newblock Sign language production using neural machine translation and generative adversarial networks.
\newblock In \emph{Proceedings of the 29th British Machine Vision Conference (BMVC 2018)}. British Machine Vision Association, 2018.

\bibitem[Stoll et~al.(2022)Stoll, Mustafa, and Guillemaut]{stoll2022there}
Stephanie Stoll, Armin Mustafa, and Jean-Yves Guillemaut.
\newblock There and back again: 3d sign language generation from text using back-translation.
\newblock In \emph{2022 International Conference on 3D Vision (3DV)}, pages 187--196. IEEE, 2022.

\bibitem[Sutton-Spence and Woll(1999)]{sutton1999linguistics}
Rachel Sutton-Spence and Bencie Woll.
\newblock \emph{The linguistics of British Sign Language: an introduction}.
\newblock Cambridge University Press, 1999.

\bibitem[Tevet et~al.(2022)Tevet, Raab, Gordon, Shafir, Cohen-or, and Bermano]{tevet2022human}
Guy Tevet, Sigal Raab, Brian Gordon, Yoni Shafir, Daniel Cohen-or, and Amit~Haim Bermano.
\newblock Human motion diffusion model.
\newblock In \emph{The Eleventh International Conference on Learning Representations}, 2022.

\bibitem[Tiwari et~al.(2022)Tiwari, Antic, Lenssen, Sarafianos, Tung, and Pons-Moll]{tiwari22posendf}
Garvita Tiwari, Dimitrije Antic, Jan~Eric Lenssen, Nikolaos Sarafianos, Tony Tung, and Gerard Pons-Moll.
\newblock Pose-ndf: Modeling human pose manifolds with neural distance fields.
\newblock In \emph{European Conference on Computer Vision ({ECCV})}, 2022.

\bibitem[Viitaniemi et~al.(2014)Viitaniemi, Jantunen, Savolainen, Karppa, and Laaksonen]{viitaniemi-etal-2014-pot}
Ville Viitaniemi, Tommi Jantunen, Leena Savolainen, Matti Karppa, and Jorma Laaksonen.
\newblock {S}-pot - a benchmark in spotting signs within continuous signing.
\newblock In \emph{Proceedings of the Ninth International Conference on Language Resources and Evaluation ({LREC}'14)}, pages 1892--1897, Reykjavik, Iceland, 2014. European Language Resources Association (ELRA).

\bibitem[Von~Agris et~al.(2008)Von~Agris, Knorr, and Kraiss]{von2008significance}
Ulrich Von~Agris, Moritz Knorr, and Karl-Friedrich Kraiss.
\newblock The significance of facial features for automatic sign language recognition.
\newblock In \emph{2008 8th IEEE international conference on automatic face \& gesture recognition}, pages 1--6. IEEE, 2008.

\bibitem[Wilbur and Kak(2006)]{wilbur2006purdue}
Ronnie Wilbur and Avinash~C Kak.
\newblock Purdue rvl-slll american sign language database.
\newblock 2006.

\bibitem[Zahedi et~al.(2005)Zahedi, Keysers, Deselaers, and Ney]{zahedi2005combination}
Morteza Zahedi, Daniel Keysers, Thomas Deselaers, and Hermann Ney.
\newblock Combination of tangent distance and an image distortion model for appearance-based sign language recognition.
\newblock In \emph{Pattern Recognition: 27th DAGM Symposium, Vienna, Austria, August 31-September 2, 2005. Proceedings 27}, pages 401--408. Springer, 2005.

\bibitem[Zelinka et~al.(2019)Zelinka, Kanis, and Salajka]{zelinka2019nn}
Jan Zelinka, Jakub Kanis, and Petr Salajka.
\newblock Nn-based czech sign language synthesis.
\newblock In \emph{Speech and Computer: 21st International Conference, SPECOM 2019, Istanbul, Turkey, August 20--25, 2019, Proceedings 21}, pages 559--568. Springer, 2019.

\bibitem[Zhang et~al.(2023)Zhang, Tian, Zhang, Li, An, Sun, and Liu]{pymafx2023}
Hongwen Zhang, Yating Tian, Yuxiang Zhang, Mengcheng Li, Liang An, Zhenan Sun, and Yebin Liu.
\newblock Pymaf-x: Towards well-aligned full-body model regression from monocular images.
\newblock \emph{IEEE Transactions on Pattern Analysis and Machine Intelligence}, 2023.

\bibitem[Zhou et~al.(2023)Zhou, Chen, Clap{\'e}s, Wan, Liang, Escalera, Lei, and Zhang]{zhou2023gloss}
Benjia Zhou, Zhigang Chen, Albert Clap{\'e}s, Jun Wan, Yanyan Liang, Sergio Escalera, Zhen Lei, and Du Zhang.
\newblock Gloss-free sign language translation: Improving from visual-language pretraining.
\newblock In \emph{Proceedings of the IEEE/CVF International Conference on Computer Vision}, pages 20871--20881, 2023.

\end{thebibliography}
